%% file: _20_IROS.tex
\documentclass[letterpaper, 10 pt, conference]{ieeeconf}  % Comment this line out if you need a4paper

\IEEEoverridecommandlockouts                              % This command is only needed if 
\usepackage{graphics} % for pdf, bitmapped graphics files
\usepackage[T1]{fontenc}
\usepackage{amsmath} % assumes amsmath package installed
\usepackage{amssymb}  % assumes amsmath package installed
\usepackage{hyperref}
\hypersetup{
    colorlinks=true,
    linkcolor=blue,
    filecolor=magenta,      
    urlcolor=blue,
}

\usepackage{color, colortbl}
\definecolor{LightCyan}{rgb}{0.88,1,1}
\usepackage{multirow}
\usepackage{siunitx}
\DeclareSIUnit\RPM{rpm}

\usepackage{xcolor}
\usepackage{amssymb}
\usepackage{tikz}
\usetikzlibrary{shapes,arrows}
\usepackage{verbatim}
\usetikzlibrary{positioning}
\usepackage[normalem]{ulem} % [normalem] removes the underlines in the reference section.

\usepackage{algorithm}
\usepackage[end]{algpseudocode} % to remove if/while ends type: [noend]
\usepackage{etoolbox}
\newcommand{\guillaume}[1]{{\textcolor{orange}{[\textbf{Guillaume : #1}]}}}

\title{\LARGE \bf Model Predictive Path Integral Control Framework for Partially Observable Navigation: A Quadrotor Case Study}

\author{Ihab S. Mohamed$^{1}$ and Guillaume Allibert$^{2}$ and Philippe Martinet$^{1}$% <-this % stops a space
\thanks{This research was supported by the ANR CLARA project (ANR-18-CE33-0004).}% <-this % stops a space
\thanks{$^{1}$Ihab S. Mohamed and Philippe Martinet are with the Universit\'e C\^{o}te d'Azur, Inria, France,
        {\tt\small \{ihab.mohamed, philippe.martinet\}@inria.fr}}%
\thanks{$^{2}$Guillaume Allibert is with the Universit\'e C\^{o}te d'Azur, CNRS, I3S, France. {\tt\small allibert@unice.fr}}%
}

\begin{document}

\maketitle
\thispagestyle{empty}
\pagestyle{empty}

%%%%%%%%%%%%%%%%%%%%%%%%%%%%%%%%%%%%%%%%%%%%%%%%%%%%%%%%%%%%%%%%%%%%%%%%%%%%%%%%
\begin{abstract}
Recently, Model Predictive Path Integral (MPPI) control algorithm has been extensively applied to autonomous navigation tasks, where the cost map is mostly assumed to be known and the 2D navigation tasks are only performed. In this paper, we propose a generic MPPI control framework that can be used for 2D or 3D autonomous navigation tasks in either fully or partially observable environments, which are the most prevalent in robotics applications. This framework exploits directly the 3D-voxel grid acquired from an on-board sensing system for performing collision-free navigation. We test the framework, in realistic RotorS-based simulation, on goal-oriented quadrotor navigation tasks in a cluttered environment, for both fully and partially observable scenarios. Preliminary results demonstrate that the proposed framework works perfectly, under partial observability, in 2D and 3D cluttered environments. 
%\textcolor{red}{Supplementary video is available at \url{https://project.inria.fr/chorale/}}.

\end{abstract}
%%%%%%%%%%%%%%%%%%%%%%%%%%%%%%%%%%%%%%%%%%%%%%%%%%%%%%%%%%%%%%%%%%%%%%%%%%%%%%%%
%%%%%%%%%%%%%%%%%%%%%%%%%%%%%%%%%%%%%%%%%%%%%%%%%%%%%%%%%%%%%%%%%%%%%%%%%%%%%%%%
\section*{Multimedia Material}
The supplementary video attached to this work is available at:
%\url{https://bit.ly/2PAbESO}
\url{https://urlz.fr/cs2L}
%%%%%%%%%%%%%%%%%%%%%%%%%%%%%%%%%%%%%%%%%%%%%%%%%%%%%%%%%%%%%%%%%%%%%%%%%%%%%%%%
%%%%%%%%%%%%%%%%%%%%%%%%%%%%%%%%%%%%%%%%%%%%%%%%%%%%%%%%%%%%%%%%%%%%%%%%%%%%%%%%
\section{INTRODUCTION}

Having a safe and reliable system for autonomous navigation of  robotic systems such as Unmanned Aerial Vehicles (UAVs) is a highly challenging and partially-solved problem for robotics communities, especially for cluttered and GPS-denied environments such as dense forests, crowded offices, corridors, and warehouses.
%especially for navigating autonomously in cluttered or unknown environments.
%such as dense forests, busy streets with buildings and overhanging wires, corridors or corners, and warehouses.
Such a problem is very important for solving many complex applications, such as surveillance, search-and-rescue, and environmental mapping.
To do so, UAVs should be able to navigate with complete autonomy while avoiding all kinds of obstacles in real-time. 
To this end, they must be able to (i) perceive their environment, (ii) understand the situation they are in, and (iii) react appropriately.

Obviously enough, this problem has been already addressed in the literature, particularly those works related to dynamics and control, motion planning, and trajectory generation in unstructured environments with obstacles \cite{mellinger2014trajectory, gonzalez2015review, mohta2018fast, baca2018model, ryll2019efficient}. 
Moreover, the applications of the path-integral control theory have recently become more prevalent. 
%\cite{kappen2005linear}. 
One of the most noteworthy works is Williams's iterative path integral method, namely MPPI control framework \cite{williams2017model}. In this method, the control sequence is iteratively updated to obtain the optimal solution on the basis of importance sampling of trajectories. 
In \cite{williams2018information}, authors derived a different iterative method in which the control- and noise-affine dynamics constraints, on the original MPPI framework, are eliminated. This framework is mainly based on the information-theoretic interpretation of optimal control using KL-divergence and free energy, while it was previously based on the linearization of Hamilton-Jacob Bellman (HJB) equation and application of Feynman-Kac lemma. 
Although different methods are adopted to derive the MPPI framework, they are practically equivalent and theoretically related\footnote{In the sense that the method in \cite{williams2018information} can exactly recover that in \cite{williams2017model} if dynamics is considered to be affine in control. In other words, the iterative method in \cite{williams2018information} can be seen as the generalization of the latter method.}.
An extension to Williams's information-theoretic-based approach is presented in \cite{kusumoto2019informed}, where a learned model is used to generate informed sampling distributions.
%that imitate samples generated from the classical approach. In \cite{okada2017path}, Okada proposed so-called Path Integral Networks (PI-Net) to mimic the iterative path integral method.

 The attractive features of MPPI controller, over alternative methods, can be summarized as: (i) a derivative-free optimization method, i.e. no need for derivative information to find the optimal solution; (ii) no need for approximating the system dynamics and cost functions with linear and quadratic forms, i.e., non-linear and non-convex functions can be naturally employed, even that dynamics and cost models can be easily represented using neural networks; (iii) planning and execution steps are combined into a single step, providing an elegant control framework for autonomous vehicles. 
 However, one drawback of MPPI is that its convergence rate is empirically slow, which has been addressed in \cite{okada2018acceleration}.
 
In the context of autonomous navigation, it is observed that the MPPI controller has been mainly applied to the tasks of aggressive driving and UAVs navigation in 2D cluttered environments. To do so, MPPI requires a cost map, as an environment representation, to drive the autonomous vehicle. Concerning autonomous driving, the cost map is obtained either \textit{off-line} \cite{williams2018information} or from an on-board monocular camera using deep learning approaches \cite{drews2017aggressive, buyval2019model}. 
Regarding UAV navigation in cluttered environments, the obstacle (i.e., cost) map is assumed to be available, and only static 2D floor-maps are used. Conversely, in practice, the real environments are often partially observable, with dynamic obstacles. Moreover, it is noteworthy that only 2D navigation tasks are performed so far, which limits the applicability of the control framework.
For this reason, this paper focuses on MPPI for 2D and 3D navigation tasks in a previously unseen and dynamic environment. In particular, the main contributions of our work can be summarized as follows:
\begin{enumerate}
    \item We propose a generic MPPI framework for autonomous navigation in cluttered 2D and 3D environments, which are inherently uncertain and partially observable. To the best of our knowledge, this point has not been reported in the literature, which opens up new directions for research.
    \item We demonstrate this framework on a set of simulated quadrotor navigation tasks using RotorS simulator and Gazebo \cite{furrer2016rotors}, assuming that: (i) there is a priori knowledge about the environment (namely, fully observable case); (ii) there is no a priori information (namely, partially observable case). In this case, the robot is building and updating the map, which represents the environment, online as it goes along. This allows the opportunity to navigate in dynamic environments.
    \item To ensure a realistic simulation, our proposed framework is evaluated taking into account the modelling errors, noisy sensors, and windy environments.
\end{enumerate}

 This paper is organized as follows. Section \ref{Quadrotor Dynamics Model} describes the quadrotor dynamics model which represents our case study for the framework validation, whereas in Section \ref{MPPI Control Strategy} the real-time MPPI control strategy is explained. Our proposed framework is then described in Section \ref{Generic MPPI Framework} and evaluated in Section \ref{Experimental Validation and Results}. Finally, concluding remarks are provided in Section \ref{sec:conclusion}.

%%%%%%%%%%%%%%%%%%%%%%%%%%%%%%%%%%%%%%%%%%%%%%%%%%%%%%%%%%%%%%%%%%%%%%%%%%%%%%%%
%%%%%%%%%%%%%%%%%%%%%%%%%%%%%%%%%%%%%%%%%%%%%%%%%%%%%%%%%%%%%%%%%%%%%%%%%%%%%%%%

\section{Quadrotor Dynamics Model}\label{Quadrotor Dynamics Model}
\begin{figure}[!ht]
\begin{center}
\includegraphics[scale = 1]{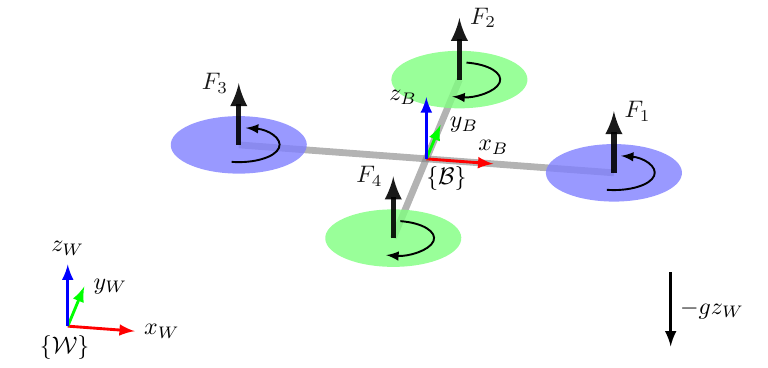}    % The printed column width is 8.4 cm.
\caption{Schematic of the considered quadrotor model in conjunction with the coordinate systems and forces acting on a vehicle frame.} 
\label{fig:quadrotorModel}
\end{center}
\end{figure}

%\subsection{Quadrotor Dynamics Model}
Considering a quadrotor vehicle model illustrated in Fig.~\ref{fig:quadrotorModel}, the dynamics model can be defined by assigning a fixed inertial frame $\mathcal{W}$ and body frame $\mathcal{B}$ attached to the vehicle. The origin of the body frame, $\mathcal{B}$, is located at the center of mass of the quadrotor, where $x_{B}$ and $y_{B}$ lie in the quadrotor plane defined by the centers of the four rotors, and $z_{B}$ is perpendicular to this plane and points upward. The inertial reference frame, $\mathcal{W}$, is defined by $x_{W}, y_{W}$, and $z_{W}$, with $z_{W}$ pointing upward. We assume that the first rotor (i.e., along the $+x_{B}$ axis) and the third rotor rotate counter clockwise (namely, $+1$), whilst the second (i.e., along the $+y_{B}$ axis) and fourth rotors rotate clockwise (namely, $-1$). 
Here, the vehicle is only subject to: (i) a gravitational acceleration $g$ in $-z_{W}$ direction; (ii) the sum of the forces generated by each rotor, $F = \sum_{i=1}^{4} F_i$, acts along the $+z_{B}$ direction.
%----------------------------
Furthermore, the Euler angles, with $ZXY$ transformation sequence, are used to model the rotation of the quadrotor in frame $\mathcal{W}$, where the roll $\phi$, pitch $\theta$, and yaw $\psi$ angles refer to a rotation about $x_{B}$, $y_{B}$, and $z_{B}$ axis, respectively. The rotation matrix from $\mathcal{B}$ to $\mathcal{W}$ is accordingly expressed as
% ----------------
\begin{equation*} \label{eq:R}
^{\mathcal{W}}\mathcal{R}_{\mathcal{B}} =\left[\begin{array}{ccc}{c_\psi c_\theta-s_\phi s_\psi s_\theta} & {-c_\phi s_\psi} & {c_\psi s_\theta+c_\theta s_\phi s_\psi} \\ {c_\theta s_\psi+c_\psi s_\phi s_\theta} & {c_\phi c_\psi} & {s_\psi s_\theta-c_\psi c_\theta s_\phi} \\ {-c_\phi s_\theta} & {s_\phi} & {c_\phi c_\theta}\end{array}\right],
\end{equation*}
where $s_{x}=\sin (x)$ and $c_{x}=\cos (x)$ $\forall$ $x \in \{\phi, \theta, \psi\}$. The transformation matrix from Euler angular velocities, $\dot{\Phi}$, to body frame angular velocity, $\Omega$, is given by
%-------------------------------------
\begin{equation*}
    T = \left[\begin{array}{ccc}{c_\theta} & {0} & {-c_\phi s_\theta} \\ {0} & {1} & {s_\phi} \\ {s_\theta} & {0} & {c_\phi c_\theta}\end{array}\right].
\end{equation*}
% -----------------------
Given all considerations above and according to the well-described models in \cite{mellinger2014trajectory} and \cite{kai2017nonlinear}, the dynamics of the position $\xi$, linear velocity $v$, orientation locally defined by Euler angles $\Phi$, and body angular rates $\Omega$ can be written as
%----------------------
\begin{equation}\label{eq:dynamics}
    \begin{aligned} 
    \dot{\xi} &=v, &m \dot{v} &= - m g e_{3}+ ^{\mathcal{W}}\mathcal{R}_{\mathcal{B}}Fe_{3}, \\ 
    \dot{\Phi} &=T^{-1} \Omega,  
     & J \dot{\Omega} &= \Gamma -\Omega \times J \Omega,
    \end{aligned}
\end{equation}
%-------------------
where $m$ is the mass of the quadrotor and $J$ its inertia matrix expressed in the body frame $\mathcal{B}$, $\xi= [x,y,z]^{T}$, $v= [v_{x},v_{y},v_{z}]^{T}$, $\Phi= [\phi, \theta, \psi]^{T}$, $\Omega= [p, q, r]^{T}$, $e_3= [0,0,1]^{T} \in \mathbb{R}^{3}$, $ F \in \mathbb{R}^{+}$ is the accumulated force (i.e., thrust) generated by all rotors and constitutes the first control input to the system, $\Gamma = [\tau_x,\tau_y,\tau_z]^{T} =[L(F_2-F_4), L(F_3-F_1), (M_1 - M_2 + M_3 -M_4)]^{T} \in \mathbb{R}^{3}$ is the total torque applied to the vehicle with its components expressed in $\mathcal{B}$ which represents the second control input, and $L \in \mathbb{R}^{+}$ is the distance from the center of the vehicle to the axis of rotation of each rotor. In this paper, the state of the system is defined as $\mathbf{x}=[x, y, z, \phi, \theta, \psi, v_{x}, v_{y}, v_{z}, p, q, r]^{T} \in \mathbb{R}^{12}$.
%-----------------------------------
Each rotor produces a vertical force, $F_i$, and moment, $M_i$, according to $F_{i}=k_{F} \omega_{i}^{2}$ and $M_{i}=k_{M} \omega_{i}^{2}$, where $\omega_i$ is the angular velocity of the $i^{th}$ rotor, $k_{F}$ is the rotor force constant, and $k_{M}$ is the rotor moment constant. Accordingly, the mapping between the control inputs, namely $F$ and $\Gamma$, and the system's input, i.e., $\omega_i$, in order to control the quadrotor, can be expressed as 
\begin{equation}\label{eq:ControlMappingEq}
\left[\begin{array}{c} F \\ \tau_x\\ \tau_y\\ \tau_z\end{array}\right]
    =\left[\begin{array}{cccc}{k_{F}} & {k_{F}} & {k_{F}} & {k_{F}} \\ {0} & {k_{F} L} & {0} & {-k_{F} L} \\ {-k_{F} L} & {0} & {k_{F} L} & {0} \\ {k_{M}} & {-k_{M}} & {k_{M}} & {-k_{M}}\end{array}\right]\left[\begin{array}{c}{\omega_{1}^{2}} \\ {\omega_{2}^{2}} \\ {\omega_{3}^{2}} \\ {\omega_{4}^{2}}\end{array}\right].
\end{equation}
%%%%%%%%%%%%%%%%%%%%%%%%%%%%%%%%%%%%%%%%%%%%%%%%%%%%%%%%%%%%%%%%%%%%%%%%%%%%%%%%
%%%%%%%%%%%%%%%%%%%%%%%%%%%%%%%%%%%%%%%%%%%%%%%%%%%%%%%%%%%%%%%%%%%%%%%%%%%%%%%%

\section{MPPI Control Strategy}\label{MPPI Control Strategy}
The MPPI controller is a stochastic Model Predictive Control (MPC) method, which can be applied to non-linear dynamics and non-convex cost objectives. So, it is a sampling-based and derivative-free approach. The key idea of MPPI is to sample thousands of trajectories, based on Monte-Carlo simulation, in real-time from the system dynamics. Each trajectory is then evaluated according to a predefined cost function. Consequently, the optimal control sequence is updated over all trajectories. This can be easily done, in real-time, by taking advantage of the parallel nature of sampling and using a Graphics Processing Unit (GPU).

Let assume that the discrete-time stochastic dynamical system has a form of 
\begin{equation}\label{eq:dynamicsEq}
\mathbf{x}_{t+1}=f\left(\mathbf{x}_{t}, \mathbf{u}_{t}+\delta \mathbf{u}_{t}\right),
\end{equation}
where $\mathbf{x}_{t} \in \mathbb{R}^{n}$ is the state vector of the system at time $t$, $\mathbf{u}_{t} \in \mathbb{R}^{m}$ denotes a control input for the system, and $\delta \mathbf{u}_{t}$ is a zero-mean Gaussian noise vector with a variance of $\Sigma_{\mathbf{u}}$, i.e., $\delta \mathbf{u}_{t} \sim \mathcal{N}(\mathbf{0}, \Sigma_{\mathbf{u}})$, which represents the control input updates. Given a finite time-horizon $t \in\{0,1,2, \cdots, T-1\}$, the objective of the stochastic optimal control problem is to find a control sequence, $\mathbf{u} = \left(\mathbf{u}_{0}, \mathbf{u}_{1}, \ldots \mathbf{u}_{T-1}\right) \in \mathbb{R}^{m \times T}$, which minimizes the expectations over all generated trajectories taken with respect to (\ref{eq:dynamicsEq}), i.e., $J= \min_{\mathbf{u}} \mathbb{E}\left[S\left(\tau\right)\right]$, where $ S\left(\tau\right) \in \mathbb{R}$ is the state-dependent cost-to-go of a trajectory $\tau=\left\{\mathbf{x}_{0}, \mathbf{u}_{0}, \mathbf{x}_{1}, \cdots, \mathbf{u}_{T-1}, \mathbf{x}_{T}\right\}$. 
Accordingly, the optimal problem can be formulated as
\begin{equation}
J=\min_{\mathbf{u}} \mathbb{E}\left[\phi\left(\mathbf{x}_T\right)+\sum_{t=0}^{T-1}\left(q\left(\mathbf{x}_{t}\right)+\frac{1}{2} \mathbf{u}_{t}^{T} R \mathbf{u}_{t}\right)\right],
\end{equation}
where $\phi\left(\mathbf{x}_T\right)$, $q\left(\mathbf{x}_{t}\right)$, and $R\in \mathbb{R}^{m \times m}$ are a final terminal cost, a state-dependent running cost, and a positive definite control weight matrix, respectively. 
To solve this optimization problem, we consider the iterative update law derived in \cite{williams2017model}, in which MPPI algorithm updates the control sequence, from $t$ onward, as
\begin{equation}
 \mathbf{u}_{t} \leftarrow \mathbf{u}_{t} +\frac{\sum_{k=1}^{K} \exp \left(-(1 / \lambda) \tilde{S}\left(\tau_{t, k}\right)\right) \delta \mathbf{u}_{t, k}}{\sum_{k=1}^{K} \exp \left(-(1 / \lambda) \tilde{S}\left(\tau_{t, k}\right)\right)}, \end{equation}
where $K$ is the number of random samples (namely, rollouts), $\lambda \in \mathbb{R}^{+}$ is a hyper-parameter so-called the inverse temperature, and $\tilde{S}\left(\tau_{t, k}\right) =\phi\left(\mathbf{x}_T\right)+\sum_{t=0}^{T-1} \tilde{q}\left(\mathbf{x}_{t}, \mathbf{u}_{t}, \delta \mathbf{u}_{t}\right)$ is the modified cost-to-go of the $k^{th}$ rollout from time $t$ onward. 
% ----------- algorithm -------------------------------
\begin{algorithm}[ht!]
\caption{Real-Time MPPI Control Scheme \cite{williams2017model}}
\label{alg:MPPIAlg.}
\hspace*{\algorithmicindent} \textbf{Given:} \\
\hspace*{1cm} $K, T$: Number of rollouts (samples) \& timesteps \\
\hspace*{1cm} $ \left(\mathbf{u}_{0}, \mathbf{u}_{1}, \ldots,
\mathbf{u}_{T-1}\right) \equiv \mathbf{u}$: Initial control sequence \\
\hspace*{1cm} $ f, \Delta t$: Dynamics \& time-step size\\
\hspace*{1cm} $\phi, q, \lambda, \nu, \Sigma_{\mathbf{u}}, R$: Cost functions/hyper-parameters \\
\hspace*{1cm} SGF: Savitzky-Galoy (SG) convolutional filter
\begin{algorithmic}[1] % [1] numbering each line
\While {task not completed} 
    %\State Generate random control variations $\delta \mathbf{u} \in \mathbb{R}^{K \times T}$
    \State $\mathbf{x}_{0} \leftarrow$ StateEstimator(), $\mathbf{x}_{0} \in \mathbb{R}^{n}$ 
    \State $ \delta \mathbf{u} \leftarrow$ RandomNoiseGenerator(), $\delta \mathbf{u} \in \mathbb{R}^{K \times T}$
    \State $\tilde{S}\left(\tau_{k}\right) \leftarrow $ TrajectoryCostInitializer(), $\tilde{S}\left(\tau_{k}\right) \in \mathbb{R}^{K}$
    \For{$k \leftarrow 0$ \textbf{to} $K-1$}
        \State $\mathbf{x} \leftarrow \mathbf{x}_{0}$
        \For{$t \leftarrow 0$ \textbf{to} $T-1$}
            \State $\mathbf{x}_{t+1} \leftarrow \mathbf{x}_{t}+ f\left(\mathbf{x}_{t}, \mathbf{u}_{t}+\delta \mathbf{u}_{t,k}\right) \Delta t$
            \State $\tilde{S}\left(\tau_{t+1, k}\right) \leftarrow \tilde{S}\left(\tau_{t, k}\right)+\tilde{q}$
        \EndFor
        \State $\tilde{S}\left(\tau_{k}\right) \leftarrow \tilde{S}\left(\tau_{t+1, k}\right)+ \phi\left(\mathbf{x}_{T}\right)$,  $\forall t=T-1$
    \EndFor
    \State $\tilde{S}_{\min} \leftarrow \min _{k}[\tilde{S}\left(\tau_{k}\right)]$
    \For{$t \leftarrow 0$ \textbf{to} $T-1$}
        \State $\mathbf{u}_{t} \leftarrow \mathbf{u}_{t}+\frac{\sum_{k=0}^{K-1} \exp \bigl( \frac{-1}{\lambda} \left[\tilde{S}\left(\tau_{t, k}\right) -\tilde{S}_{\min} \right] \bigr) \delta \mathbf{u}_{t, k}}{\sum_{k=0}^{K-1} \exp \bigl(\frac{-1}{\lambda} \left[\tilde{S}\left(\tau_{t, k}\right) -\tilde{S}_{\min} \right]\bigr)}$
    \EndFor 
    \State $\mathbf{u} \leftarrow \text{SGF}(\mathbf{u})$
    \State $\mathbf{u}_{0} \leftarrow$ SendToActuators($\mathbf{u}$)
    \For{$t \leftarrow 1$ \textbf{to} $T-1$}
        \State $\mathbf{u}_{t-1} \leftarrow \mathbf{u}_{t}$
    \EndFor
    \State $\mathbf{u}_{T-1} \leftarrow$ ControlSequenceInitializer($\mathbf{u}_{T-1}$)
    \State Check for task completion
\EndWhile
\end{algorithmic}
\end{algorithm}
%---------------------------------------------------------

In this work, the modified running cost $\tilde{q}\left(\mathbf{x}, \mathbf{u}, \delta \mathbf{u}\right)$ is defined as
\begin{equation}
\tilde{q}=q(\mathbf{x})+\frac{1}{2} \mathbf{u}^{T} R \mathbf{u}+\frac{1-\nu^{-1}}{2} \delta \mathbf{u}^{T} R \delta \mathbf{u}+\mathbf{u}^{T} R \delta \mathbf{u},    
\end{equation}

 where $\nu \in \mathbb{R}^{+}$ refers to the exploration noise which determines how aggressively the state-space is explored. It is noteworthy that the low values of $\nu$ result in the rejection of many sampled trajectories because their cost is too high, while too large values result in that the controller produces control inputs with significant chatter.

The real-time control cycle of MPPI algorithm is described in Algorithm~\ref{alg:MPPIAlg.} with more detail. 
%At each time-step $\Delta t$ and given the system dynamics, the cost function of a given task and its parameters, and an initial control sequence,
At each time-step $\Delta t$, the system current state is estimated, and a $K \times T$ random control variations are generated on a GPU using CUDA's random number generation library (lines $2:3$). Then, based on the parallel nature of sampling, all trajectory samples are executed individually in parallel. For each trajectory, the dynamics are predicted forward and its expected cost is computed (lines $5:12$), bearing in mind that the cost of each trajectory is zero-initialized (line $4$). The control sequence is then updated (lines $14:16$), taking into account the minimum sampled cost $\tilde{S}_{\min}$ (line $13$). Due to the stochastic nature of the sampling procedure which leads to significant chattering in the resulting control, the control sequence is then smoothed using a Savitzky-Galoy (SG) filter (line 17). Finally, the first control is executed (line 18), while the remaining sequence of length $T-1$ is slid down to be used at next time-step $\Delta t$ (lines $19:22$).

%%%%%%%%%%%%%%%%%%%%%%%%%%%%%%%%%%%%%%%%%%%%%%%%%%%%%%%%%%%%%%%%%%%%%%%%%%%%%%%%
%%%%%%%%%%%%%%%%%%%%%%%%%%%%%%%%%%%%%%%%%%%%%%%%%%%%%%%%%%%%%%%%%%%%%%%%%%%%%%%%

\section{Generic MPPI Framework}\label{Generic MPPI Framework}
\begin{figure}[!ht]
\begin{center}
\includegraphics[scale = 0.9]{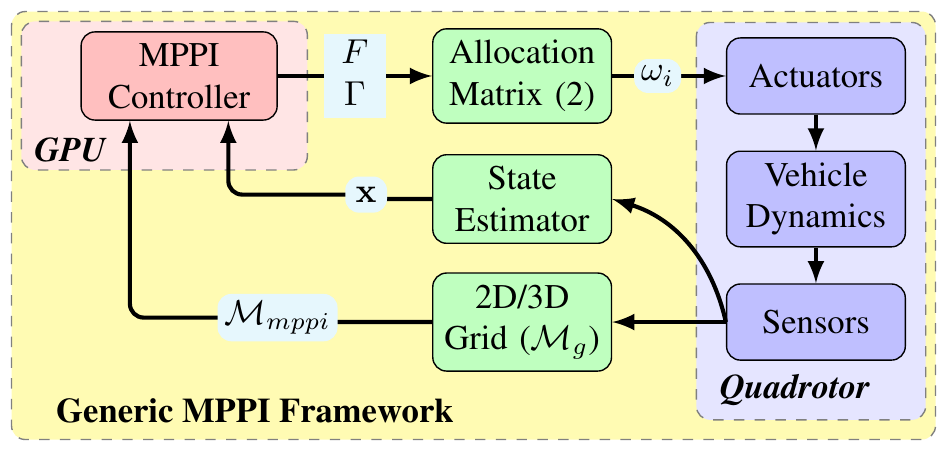}
\vspace{-0.1in}
\caption{The global architecture of our proposed framework.} 
\label{fig:ros_mppi_framework}
\end{center}
\end{figure}

In this section, we present a generic and elegant MPPI framework, as illustrated in Fig.~\ref{fig:ros_mppi_framework}, in order to not only navigate autonomously in previously unseen 2D or 3D environments while avoiding collisions with obstacles but also to explore and map them.  Moreover, we describe how the environment is represented to be used by MPPI for partially and fully observable navigation tasks. In Fig.~\ref{fig:ros_mppi_framework}, we present the block diagram of our proposed control framework integrated into the Robot Operating System (ROS). The individual components of our framework are described in detail in the following sections.
%\ref{Experimental Validation and Results}}.  
%-----------------------------------------------------------------------
% --------------------------------------------------------------------------------
\subsection{Environment Representation}
\begin{figure}[!ht]
\vspace{-0.26in}
\begin{center}
\input{layers.tikz}
\vspace{-0.2in}
\caption{3D environment representation of a generic MPPI.} 
\label{fig:layers}
\end{center}
\end{figure}
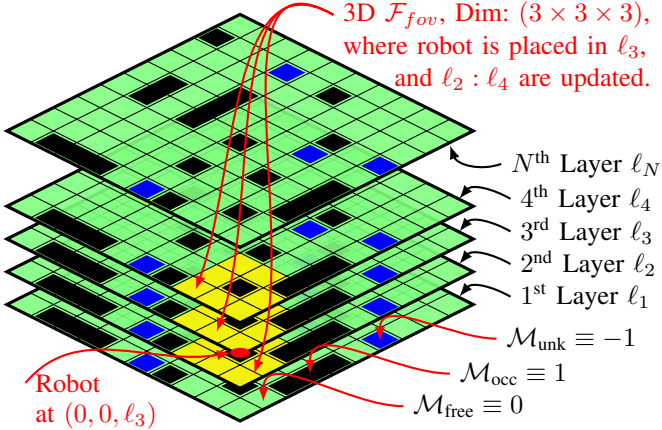

To clarify how MPPI can be used for 2D or 3D navigation in unseen environments, we assume that the environment is discretized into a 2D or 3D grid $\mathcal{M}_g$ (see Fig.~\ref{fig:ros_mppi_framework}), where each cell is labeled as free, occupied, or unknown, i.e., $\mathcal{M}_g =\mathcal{M}_{\text{free}} \cup \mathcal{M}_{\text{occ}} \cup \mathcal{M}_{\text{unk}}$. 
In practice, this labeling can be acquired from depth sensors using OctoMap \cite{hornung2013octomap}. 
As the 3D grid is a grid of cubic volumes of equal size called voxels, we can represent the environment by a set of layers $\ell_{N}$ along $z_{W}$ direction, where $\ell_{N} =(\frac{\ell_z}{r})$, $\ell_z$ refers to the real environment's height, and $r$ is the voxel size. Accordingly, each layer represents a 2D occupancy grid, producing in-total $\ell_{N}$ 2D grids for a given environment, as illustrated in Fig.~\ref{fig:layers}.
Since this work is mainly concerned with the control framework, the perception problem is not presently covered. 
Thus, the perception is here imitated by the so-called 2D or 3D mask $\mathcal{F}_{fov}$, which represents the sensor's field of view (FoV) as the robot cannot normally perceive the entire environment. 
% ------------------------------------------------
For the sake of simplicity, we assume that the 3D $\mathcal{F}_{fov}$ has dimensions $f_{x} \times f_{y} \times f_{z}$ and an orientation $f_{\theta}$, where $f_{x}$ and $f_{y}$ represent the number of cells in the 2D grid along $x_{W}$ and $y_{W}$ axes. While $f_{z}$ represents the number of layers along $z_{W}$, assuming that the robot's vertical FoV $f_N$ is between $[(\textit{int}(\frac{-f_{z}}{2})+\ell_n):(\textit{int}(\frac{f_{z}}{2})+\ell_n)]$ where the robot is located in layer $\ell_n$ (see Fig.~\ref{fig:layers}). Note that the 3D $\mathcal{F}_{fov}$ is equivalent to $f_N$ 2D $\mathcal{F}_{fov}$.
%-----------------------------------------------------------------------------
\subsection{Fully/Partially Observable 2D/3D Navigation Tasks}
Let assume that $\mathcal{M}_g$ represents the global map representation of a given environment. This map is initialized using a priori knowledge about the environment. Let $\mathcal{M}_{mppi}$ be the local map of MPPI, which has the same size as $\mathcal{M}_g$ and each cell is initialized with $-1$ referring to unknown cells. This local map is continuously updated using an onboard sensing system, as depicted in Fig.~\ref{fig:ros_mppi_framework}.
%and thus more information is captured, as depicted in Fig.~\ref{fig:ros_mppi_framework}.} 
%-----------------------------------------
\subsubsection{Fully Observable Case} 
In a fully observable case, MPPI is directly provided with a global map $\mathcal{M}_g$ to compute the trajectories' cost for avoiding obstacles, taking into account the current robot position obtained from a state estimator. For a 2D navigation task, a 2D floor grid is sufficient; only the first layer, $\ell_{1}$, of 3D voxel-grid will be accessed by MPPI.
While, for the 3D case, the 3D voxel-grid is required for performing collision-free navigation. 
Clearly, given the robot's position, particularly its $z$-component, the corresponding layer, $\ell_{n}$, to $z$-component is used for evaluating the cost-to-go of each sampled trajectory (see Fig.~\ref{fig:layers}). At the moment, the main limitation of the fully observable case is that our framework is only able to handle static environments; this is an important motivation for defining a partially observable navigation task that is prevalent in robotics applications. 
%\guillaume{remark on ``our''}
%-----------------------------------------
\subsubsection{Partially Observable Case} 
In this case, the robot is located in a previously unseen and dynamic environment and must navigate to: (i) a predefined goal, (ii) or explore and map that area. As the environment is unknown and $\mathcal{M}_g$ is not directly accessible by MPPI, the robot must perceive its environment (in our case, through the predefined mask $\mathcal{F}_{fov}$) and react appropriately. Here, the MPPI map $\mathcal{M}_{mppi}$, which is initialized with $-1$, is fed directly into the control framework as a local map for avoiding obstacles. This map is continuously updated as the robot moves around. For the sake of clarity, let us consider the map in Fig.~\ref{fig:layers} as $\mathcal{M}_g$, where the robot is located in $(0,0, \ell_3)$ and $\mathcal{F}_{fov}$ has dimensions of $3 \times 3 \times 3$. As a consequence, based on the intersection between $\mathcal{M}_g$ and $\mathcal{F}_{fov}$, the layers from $\ell_2$ to $\ell_4$ in $\mathcal{M}_{mppi}$ are updated, while other layers have remained constant. 
% ----------- algorithm -------------------------------
\begin{algorithm}[ht!]
\caption{Real-Time Generic MPPI Framework}
\label{alg:GenericMPPIAlg.}
\hspace*{\algorithmicindent} \textbf{Given:} \\
\hspace*{1cm} $\mathcal{M}_g, \mathcal{M}_{mppi}$: Global and local map of MPPI \\
\hspace*{1cm} $\mathcal{F}_{fov}$: Sensor's FoV and its parameters
\begin{algorithmic}[1] % [1] numbering each line
\While {task not completed} 
    \State $ \mathbf{x}_{0} \leftarrow$ StateEstimator(), $\mathbf{x}_{0} \in \mathbb{R}^{n}$ 
    \State $ \mathcal{M}_{mppi} \leftarrow$ MapUpdate($\mathcal{M}_g \cap \mathcal{F}_{fov}$)
    \State $\mathbf{u}_{0} \leftarrow$ MPPIController($\mathcal{M}_{mppi}$)
    \State Check for task completion
\EndWhile
\end{algorithmic}
\end{algorithm}

Generally speaking, the real-time implementation of our proposed generic MPPI framework for 2D or 3D navigation in cluttered environments is better described in Algorithm~\ref{alg:GenericMPPIAlg.}, which employs the MPPI control scheme described above. At each time-step $\Delta t$, the current state is estimated (line 2). The local map of MPPI $\mathcal{M}_{mppi}$ is then updated accordingly, given the global map $\mathcal{M}_g$ (line 3). As discussed previously, $\mathcal{M}_g$ in conjunction with $\mathcal{F}_{fov}$ is used for updating the local map $\mathcal{M}_{mppi}$, as the perception modules have not been considered in the current work. Thus, in practice, the robot must be equipped with an on-board sensing system, e.g. depth camera or laser scanner, with a maximum FoV $\mathcal{F}_{fov}$. This currently sensed data is used to obtain a 2D or 3D occupancy map, which is continuously updated, i.e., $\mathcal{M}_{mppi}$. For instance, a quadrotor-based exploration algorithm is proposed in \cite{cieslewski2017rapid} to build a real-time 3D map. Finally, this map enables the controller to find the optimal control $\mathbf{u}_{0}$ to be executed, resulting in collision-free navigation (line 4).

%%%%%%%%%%%%%%%%%%%%%%%%%%%%%%%%%%%%%%%%%%%%%%%%%%%%%%%%%%%%%%%%%%%%%%%%%%%%%%%%
%%%%%%%%%%%%%%%%%%%%%%%%%%%%%%%%%%%%%%%%%%%%%%%%%%%%%%%%%%%%%%%%%%%%%%%%%%%%%%%%
\section{Simulation Details and Results}\label{Experimental Validation and Results} 
In this section, we describe how we evaluate our approach in terms of simulation scenarios and performance metrics. Moreover, we conduct realistic simulations to evaluate and demonstrate the performance of the proposed framework.
\subsection{Simulation Setup, Scenarios, and Metrics}
%--------------------------------------------------------
\subsubsection{Simulation Setup} In order to evaluate the performance of our proposed MPPI framework in a previously unseen and cluttered environment, simulation studies have been performed using RotorS simulator and Gazebo \cite{furrer2016rotors}. 
%\guillaume{The choice of this software is justified by the fact that it allows simulating the dynamic behavior of the most faithful UAV of reality by allowing to add measurement noises on the sensors, external disturbances such as wind, etc...'' }
The parameters of the real simulated quadrotor (namely, Humminbird quadrotor) are tabulated in Table~\ref{table:SysParameters}.
To ensure a realistic RotorS-based simulation, all navigation tasks are carried out by (i) using noisy sensors such as GPS and IMU, (ii) adding external disturbances such as continuous wind with changing speed and direction, as proposed in \cite{du2019boarr}, and (iii) considering $\pm$10\% of modelling errors in the real values of mass $m$ and inertia $J$ of the prediction model given in (\ref{eq:dynamics}).
%\guillaume{To make realistic simulation, in all presented simulations, we make use of the “erroneous” estimated parameters instead of the real values for mass and inertia: $\hat{m}=23.543kg$, $\hat{J}=diag(1,2,3)$ which correspond approximately to errors around 10\%. Moreover, noie sensors and wind with the following characteristics .... are also added.}
The MPPI controller has a time horizon $t_p$ of \SI{3}{\second}, a control frequency of \SI{50}{\hertz}, and generates 2700 samples each time-step $\Delta t$. The rest of its hyper-parameters are also listed in Table~\ref{table:SysParameters}, where the $2.5$ value in $\Sigma_{\mathbf{u}}$ represents the noise in the thrust input $F$ and $R= \lambda \Sigma_{\mathbf{u}}^{-1}$. For the SG filter, we set the length of filter window and order of the polynomial function to 51 and 3, respectively. The real-time execution of MPPI is performed on a GeForce GTX 1080 Ti, where all algorithms are written in Python and have been implemented using ROS.
%----------------------------------------------------------------------------------
\begin{comment}
The simulated quadrotor, i.e., Hummingbird UAV, has the following parameters: $m= \SI{0.716}{\kilo\gram}$, $L= \SI{0.17}{\metre}$, $J=\operatorname{Diag}\left(\num{7d-3}, \num{7d-3}, \num{12d-3}\right)$\,\si{\kilo\gram\metre\squared}, $g = \SI{9.81}{\metre\per\second\squared}$, $k_{F}=$ \SI{8.55d-6}{\newton\per\RPM\squared}, and $k_{M}=$ \SI{1.6d-2}{\newton\per\RPM\squared}. 
The MPPI controller has a time horizon $t_p$ of \SI{3}{\second}, a control frequency of \SI{50}{\hertz}, i.e., $T=150$, and generates 2700 samples, i.e., $K=2700$, each time-step $\Delta t$. We set the hyper-parameters of MPPI as $\lambda=0.02$, $\nu=1000$, $\Sigma_{\mathbf{u}}= \operatorname{Diag}\left(2.5, \num{5d-3}, \num{5d-3}, \num{5d-3}\right)$, where the $2.5$ value represents the noise in the thrust input $F$, and $R= \lambda \Sigma_{\mathbf{u}}^{-1} = \operatorname{Diag}\left(\num{8d-3}, 4, 4, 4\right)$. For the SG filter, we set the length of filter window and order of the polynomial function to 51 and 3, respectively. 
The real-time execution of MPPI is performed on a GeForce GTX 1080 Ti GPU, where all algorithms are written in Python and have been implemented using the Robot Operating System (ROS).
\end{comment}
%-----------------------------------------------------------------------------------
\begin{table}[!ht]
\caption{Parameters of quadrotor and MPPI}
\centering
\begin{tabular}{|l| l||l| l|} 
\hline
 Parameter  & Value                         & Parameter     & Value\\
 \hline  
 \hline
 $m$ [\si{\kilo\gram}]       & \num{0.716}               & $t_p$ [\si{\second}]           & \num{3}\\
 $L$  [\si{\metre}]       & \num{0.17}                    & $T$            & \num{150}\\  
 $g$  [\si{\metre\per\second\squared}]       & \num{9.81}  & $K$            & \num{2700}\\
 $k_{F}$ [\si{\newton\per\RPM\squared}]    & \num{8.55d-6} & $\lambda$            & \num{0.02}\\
 $k_{M}$ [\si{\newton\per\RPM\squared}]    & \num{1.6d-2}  & $\nu$            & \num{1000}\\
 \hline \hline
 $J$ [\si{\kilo\gram\;\metre\squared}]      & \multicolumn{3}{l|}{$\operatorname{Diag}\left(\num{7d-3}, \num{7d-3}, \num{12d-3}\right)$}\\
 $\Sigma_{\mathbf{u}}$                      & \multicolumn{3}{l|}{$\operatorname{Diag}\left(2.5, \num{5d-3}, \num{5d-3}, \num{5d-3}\right)$}\\
 $R$                                        & \multicolumn{3}{l|}{$\operatorname{Diag}\left(\num{8d-3}, 4, 4, 4\right)$}\\
\hline
\end{tabular}
\label{table:SysParameters}
\end{table}
%\textcolor{red}{Figure~\ref{fig:ros_mppi_framework} shows the block diagram of our generic control framework integrated into ROS.} 
~\\In our RotorS-based simulations, the full-state information, $\mathbf{x}=[x, y, z, \phi, \theta, \psi, v_{x},$ $v_{y}, v_{z}, p, q, r]^{T}$, is directly provided, using an \textit{odometry} estimator based on an Extended Kalman Filter (EKF), as an input to the controller. While, since the actual control signal of the quadrotor is the angular velocity of each rotor $\omega_i$, the controller outputs $F$ and $\Gamma$ are directly converted into $\omega_i$ using~(\ref{eq:ControlMappingEq}), to be sent to the quadrotor as \textit{Actuators} message. In summary, the closed-loop of MPPI in ROS, at each $\Delta t$, can be summarized as: (i) MPPI first receives the \textit{odometry} message and the updated map $\mathcal{M}_{mppi}$ obtained from the on-board sensor; (ii) the control action is then computed and published. 
As mentioned before, the whole process of our proposed framework is summarized in Fig.~\ref{fig:ros_mppi_framework}.
%Therefore, it is important to guarantee that the \textit{odometry} message is published at the desired control rate, while the 3D map can be published at a lower rate.
%---------------------------------------------------------------------------------------------
\begin{comment}
Since we are interested in goal-oriented quadrotor navigation tasks in cluttered environments, the state-dependent cost function is defined as $q(\mathbf{x})=(\mathbf{x}-\mathbf{x}^{\text{des}})^{\mathrm{T}} Q (\mathbf{x}-\mathbf{x}^{\text{des}})+ \num{d8}C_1+ \num{d5}C_2$, where $C_1$ indicates the collision with ground or obstacles, i.e., 
$C_{1}=\{\mathbf{x}\,|\,(z<0) \wedge (\mathcal{M}_{mppi}[x,y,z]=1)\}$, while $C_{2}=\{\mathbf{x}\,|\,(\lvert v_{x} \wedge v_{y} \wedge v_{z} \lvert > v_\text{max}) \wedge \textcolor{red}{(\lvert \cos{\phi}\lvert \wedge \lvert\cos{\theta} \lvert < 0.1)} \wedge (z > 8.5)\}$ which prevents the quadrotor from (i) going too fast, (ii) using too aggressive \textcolor{red}{roll and pitch} angles, or (iii) colliding with the ceiling. $\mathbf{x}^\text{des}$ refers to the desired position to be reached and its orientation, where  
$\mathbf{x}^\text{des}=(x^\text{des},y^\text{des},z^\text{des},0,0,\psi^\text{des},\operatorname{zeros}(1,6))$. The weighing matrix $Q$ is set to $\operatorname{Diag}(2.5,2.5,5,1,1,50,\operatorname{zeros}(1,6))\,\forall$ $v_\text{max} \le \SI{1.5}{\metre\per\second}$, otherwise $Q=(5,5,15,30,30,50,\operatorname{zeros}(1,6))$. 
\end{comment}
%---------------------------------------------------------------------------------------------------------

Since we are interested in goal-oriented quadrotor navigation tasks in cluttered environments, the state-dependent cost function is defined as
\begin{equation*}
q(\mathbf{x})=(\mathbf{x}-\mathbf{x}^{\text{des}})^{\mathrm{T}} Q (\mathbf{x}-\mathbf{x}^{\text{des}})+ \num{d8}C_1+ \num{d5}C_2,   
\end{equation*}
where:
\begin{equation*}
\begin{aligned}
C_{1}&=\{\mathbf{x}:\,(z<0) \,\textbf{or}\, (\mathcal{M}_{mppi}[x,y,z]=1)\}, \\
C_{2}&=\{\mathbf{x}:\,(\|v\| > v_\text{max}) \,\textbf{or}\, \left(\left((\lvert \cos{\phi}\lvert\right) \,\textbf{or}\, \left( \lvert\cos{\theta} \lvert\right)\right) < 0.1)\\
& \hspace*{1.24in} \textbf{or}\, (z > 8.5)\},\\
\mathbf{x}^\text{des}&=(x^\text{des},y^\text{des},z^\text{des},0,0,\psi^\text{des},\operatorname{zeros}(1,6)),\\
Q\!\!&=\!\!\left\{\begin{array}{l}{\!\!\!\!\operatorname{Diag}(2.5,2.5,5,1,1,50,\operatorname{zeros}(1,6))}, \, \forall v_\text{max} \le 1.5 \frac{\text{m}}{\text{s}},\\ 
{\!\!\!\!\operatorname{Diag}(5,5,15,30,30,50,\operatorname{zeros}(1,6)), \;  \text{Otherwise}.}\end{array}\right.
\end{aligned}
\end{equation*}
The first term $C_1$ indicates the collision with ground or obstacles, while $C_{2}$ prevents the quadrotor from (i) going too fast, (ii) using too aggressive roll and pitch angles, or (iii) colliding with the ceiling, where $C_1$ and $C_2$ are boolean variables. $\mathbf{x}^\text{des}$ refers to the desired position to be reached and its orientation, while $Q$ is a weighing matrix. 
% -----------------------------------------------------------------------
\subsubsection{Simulation Scenarios}
\begin{figure}[ht!]
\begin{center}
\includegraphics[height= 1.5in, width= 3.1in]{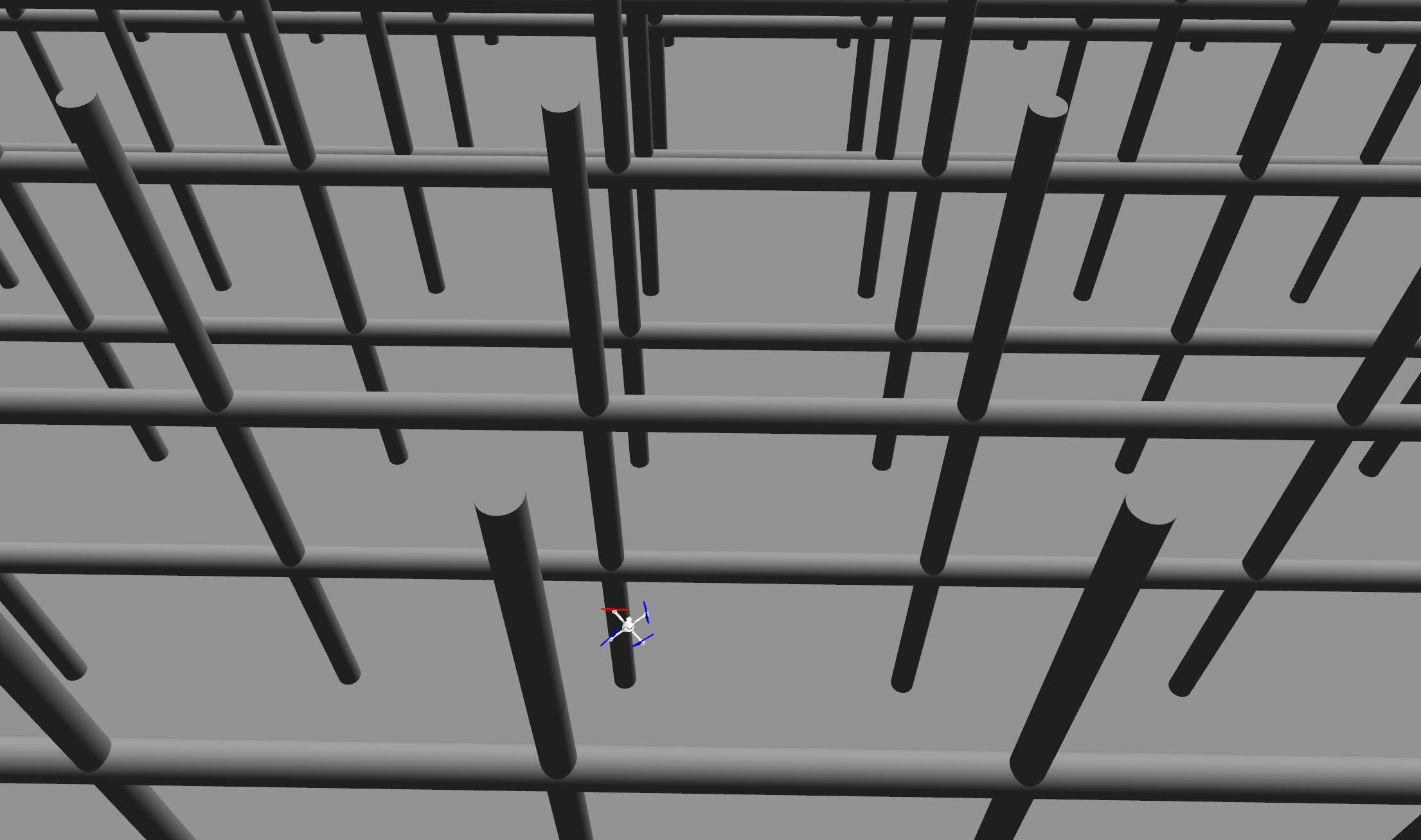} 
\caption{The 3D cluttered environment used for 3D tasks.} 
\label{fig:3Dmap_gazebo}
\end{center}
\end{figure}

Two different scenarios are considered for evaluating the proposed framework: \textit{2D scenario} and \textit{3D scenario}. 
The \textit{2D scenario} refers to a $40 \times 40\times 8.5$\,\si{\metre} windy forest of cylindrical obstacles placed in a 2D grid pattern, where each cylinder has a radius of \SI{0.16}{\metre} with equal spacing of \SI{4}{\metre} apart. 
The \textit{3D scenario} refers to the same environment described in the \textit{2D scenario}, while we have added two horizontal layers of cylinders at $z =$~\SI{3}{\metre} and \SI{6}{\metre} with the same spacing each, as shown in Fig.~\ref{fig:3Dmap_gazebo}. The voxel size $r$ is set to \SI{0.2}{\metre}, to ensure high accuracy and to meet the reality when a 3D occupancy grid is involved. As a result, the 3D voxel grid $\mathcal{M}_g$ has 43 layers, each layer represents a $40 \times 40$ 2D grid.\footnote{The cell size of 2D grid is \SI{1}{\metre} (i.e., 1:1 scale), since all obstacles, in our case, are placed in a 2D grid pattern with integer numbers.}
The \textit{3D scenario} is used for performing fully and partially observable 3D navigation tasks, while the former is used for the 2D tasks. 
For both scenarios, two different cases are considered: (i) Fully Observable Case (\textit{FOC}), in which an a priori knowledge about the environment is used to initialize the global map $\mathcal{M}_{g}$; in \textit{FOC}, $\mathcal{M}_{mppi}$ is exactly  $\mathcal{M}_{g}$; and (ii) Partially Observable Case (\textit{POC}), where it is assumed that there is no a priori information about the environment; for this reason, the local map $\mathcal{M}_{mppi}$ is discovered and built online as the robot moves around.
%-----------------------------------------------
% ----------------------------------------------------------------
\subsubsection{Performance Metrics} We define two metrics for evaluation. First, the performance of MPPI in \textit{POC} is compared to its performance in \textit{FOC}, considering the latter case as a baseline. To achieve a fair comparison, in all simulations, the robot navigates to the same specified goals with a maximum velocity $v_\text{max}$ of \SI{1.5}{\metre\per\second}. The predefined goals (in \si{[\metre]}) in order are: $G_1= (23,38,1.5), G_2= (40,23,4), G_3= (22,0,8), G_4= (0,22,5)$, then the UAV will land. While, at each $\Delta t$, the desired yaw angle $\psi^\text{des}$ is updated, letting the front camera points towards the next goal. 
For both cases (\textit{FOC}/\textit{POC}), we use a number of indicators, as a second metric, to describe the general performance such as the number of collisions $N_{\text{col}}$, task completion percentage $t_{\text{comp}}$, average completion time $t_{\text{av}}$, average flying distance $d_{\text{av}}$, average flying speed $v_{\text{av}}$, and average energy consumption $E_{\text{av}}$ (for more details, we refer to \cite{du2019boarr}). In all simulations, the task is considered to be terminated if the quadrotor reached the given goals (i.e., $t_{\text{comp}}=100\%$) or crashed into an obstacle (i.e., $N_{\text{col}}=1$). 
% ----------------------------------------------------------------
\subsection{Simulation Results} The performance of our proposed framework is validated for both fully and partially observable 2D/3D quadrotor navigation tasks, considering the predefined goals. In all partially observable tasks, we set $\mathcal{F}_{fov}$ to $5 \times 5 \times 3$ \si{\metre}, while $f_{\theta}$ represents the angle between the current and next goal. 
To test whether the quadrotor is able to navigate successfully through the cluttered environment, we performed 5 trials for both \textit{FOC} and \textit{POC}. The reader is invited to watch the whole simulation results at \url{https://urlz.fr/cs2L}. 
% --------------------------------------------------------------------------------
\subsubsection{2D Navigation Results} 
\begin{figure}[ht!]
%\vspace*{-0.1in}
\begin{center}
\hspace*{-0.3in}\includegraphics[scale = 0.6]{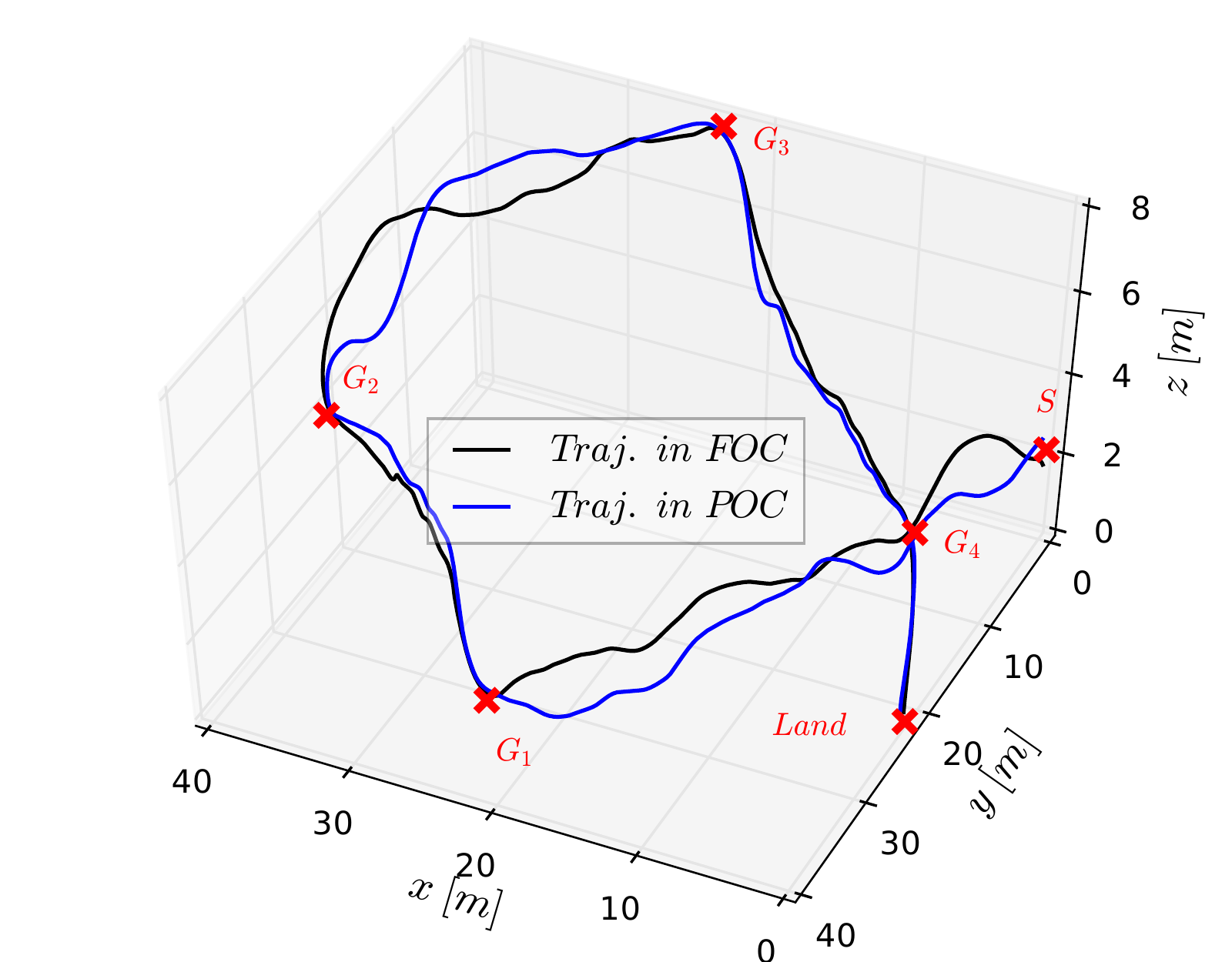}
\vspace{-0.1in}
\caption{Comparison of 3D trajectories in a \textit{2D scenario}.} 
\label{fig:3DTraj_2d}
\end{center}
\end{figure}

\begin{figure}[ht!]
%\vspace*{-0.1in}
\begin{center}
\includegraphics[scale = 0.51]{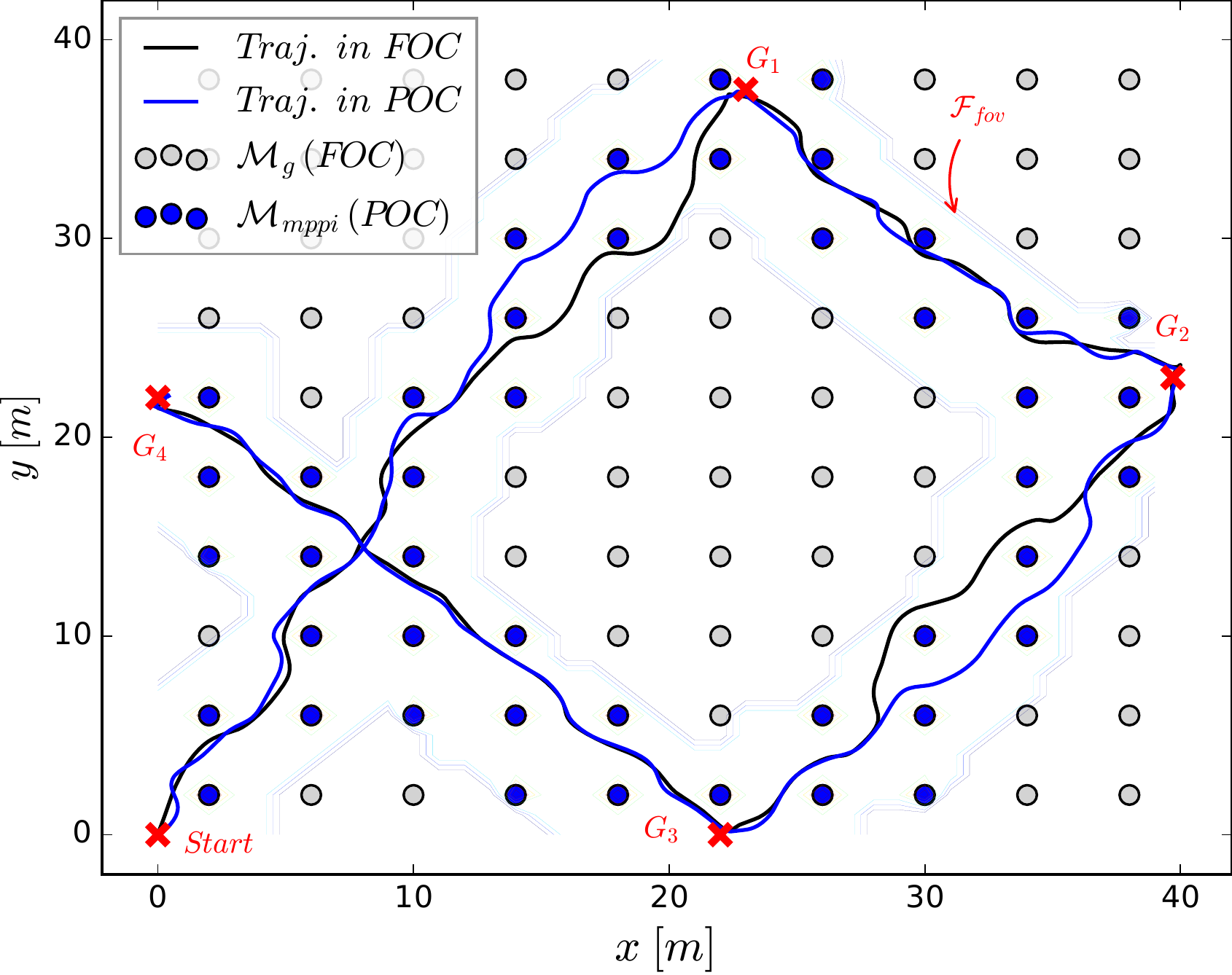} 
\vspace{-0.1in}
\caption{The \SI{40}{\metre} obstacle field map, i.e., 2D $\mathcal{M}_g$, and its updated 2D $\mathcal{M}_{mppi}$ map using a $5 \times 5$ 2D $\mathcal{F}_{fov}$. Two lines represent the trajectories generated by MPPI in both cases, i.e., \textit{FOC} and \textit{POC}.} 
\label{fig:2D_MPPImap_update}
\end{center}
\end{figure}

%-----------------------------------------------
Figure~\ref{fig:3DTraj_2d} shows an example of a final trajectory generated by MPPI in the case of (i) using a 2D global map $\mathcal{M}_g$ for \textit{FOC}, or (ii) using only a 2D local map $\mathcal{M}_{mppi}$ for \textit{POC}. The 2D-floor $\mathcal{M}_g$ (with grey circles) and its updated $\mathcal{M}_{mppi}$ (with blue circles, representing the obstacles within the robot's $\mathcal{F}_{fov}$), for given goals, are shown in Fig.~\ref{fig:2D_MPPImap_update}, including the generated trajectories in both cases and the robot's FoV $\mathcal{F}_{fov}$. In both 2D \textit{FOC} and \textit{POC}, it can be seen that MPPI is able to safely navigate through the windy obstacle field, in spite of the presence of modelling errors and measurements noise.
%\vspace*{-0.3in}
%-----------------------------------------------
%----------------------------------------------
\begin{figure}[ht!]
%\vspace*{-0.2in}
\begin{center}
\hspace*{-0.7in}\includegraphics[scale = 0.6]{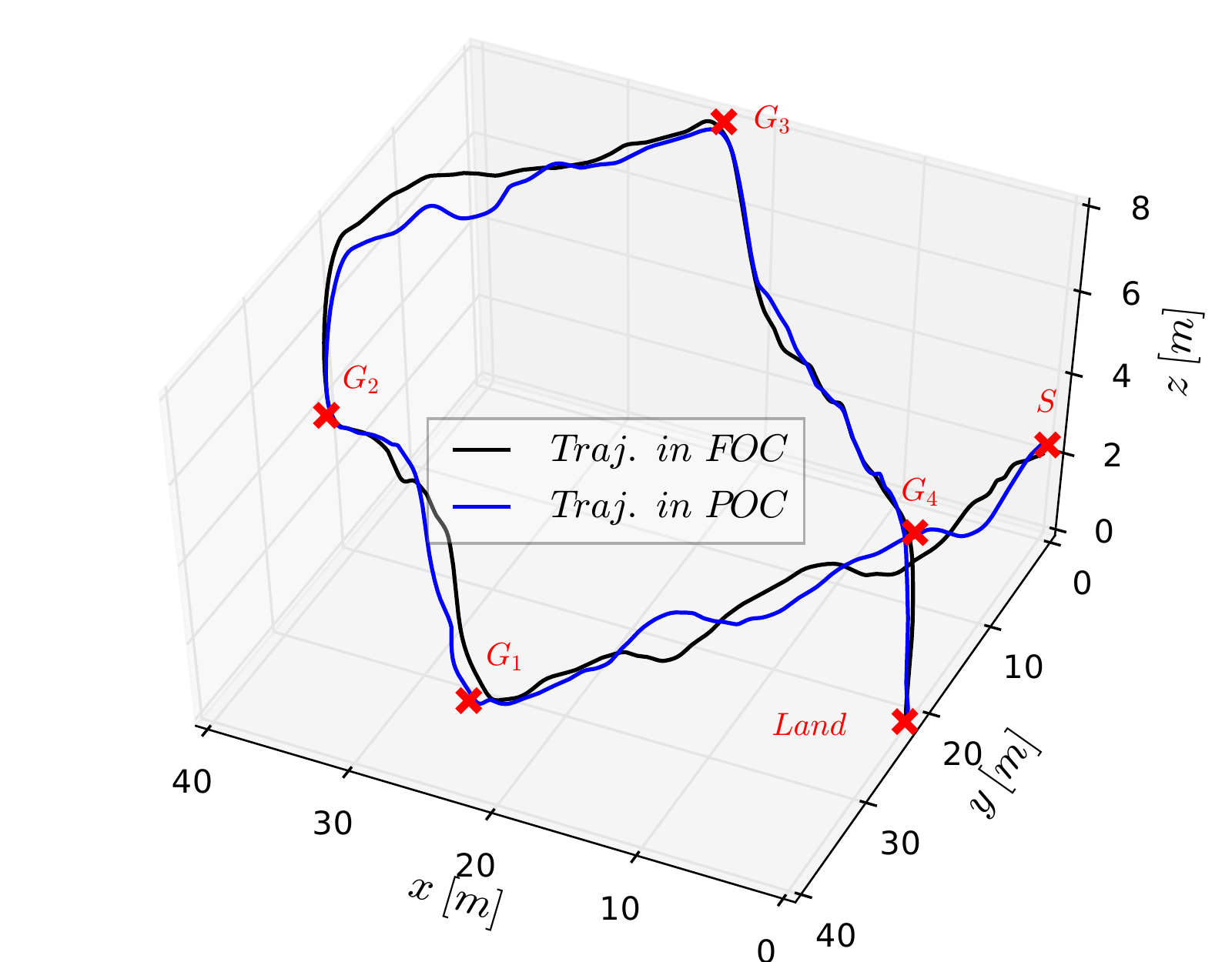}
%\vspace{-0.1in}
\caption{Comparison of 3D trajectories in a \textit{3D scenario}.} 
\label{fig:3DTraj_3d}
\end{center}
\end{figure}
%-----------------------------------------------
\subsubsection{3D Navigation Results} 
\begin{figure}[ht!]
%\vspace*{-0.1in}
\begin{center}
\hspace*{-0.1in}\includegraphics[scale = 0.51]{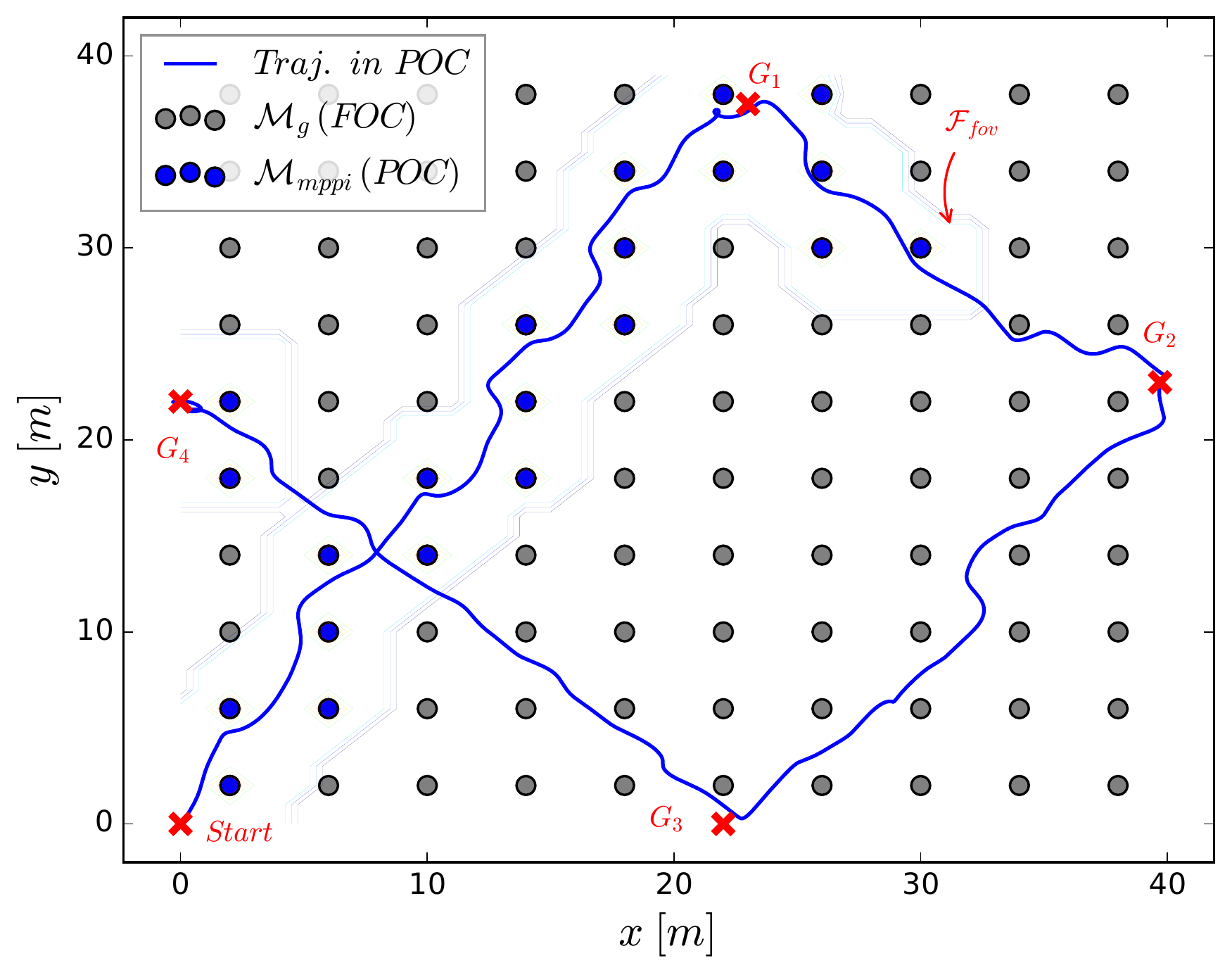}
\vspace{-0.1in}
\caption{The update of the $11^{\text{th}}$ layer $\ell_{11}$ in the 3D $\mathcal{M}_{mppi}$, showing the generated trajectory in \textit{POC}.} 
\label{fig:layer11_update}
\end{center}
\end{figure}

Another example of successfully generated trajectories for fully and partially observable 3D navigation tasks in a \textit{3D scenario} environment is depicted in Fig.~\ref{fig:3DTraj_3d}. In this scenario, since $\mathcal{M}_{mppi}$ consists of 43 2D grids, it is very difficult to visualize clearly in this paper the updates over all layers and insure that the robot performs collision-free navigation. So, first, we show only the $11^{\text{th}}$ layer $\ell_{11}$, which has been mainly updated while the robot was moving towards $G_1$ and while it was landing after task completion, as an example of how the 3D $\mathcal{M}_{mppi}$ is updated (see Fig.~\ref{fig:layer11_update}). Second, it is highly recommended to use the indicators described previously, especially for 3D cases.
%-------------------------------------------------------------------
%---------------------------------------------------------------
%The reader is invited to view a video clip showing the whole experiment at \url{https://bit.ly/2PAbESO}.\guillaume{DONE}
%-----------------------------------------------
\subsubsection{Overall Performance}
%The main results of 2D and 3D scenarios are reported in Table~\ref{table:Comparison}. 
Table~\ref{table:Comparison} shows the performance statistics for the five trials in 2D and 3D scenarios, considering the parameters of the controller tabulated in Table~\ref{table:SysParameters}.
%----------------------------------------------------------------------------------------------
%\definecolor{LightCyan}{rgb}{0.88,1,1}
\definecolor{applegreen}{rgb}{0.7, 1, 0.0}
\begin{table}[ht!]
%\caption{General performance of MPPI for fully and partially observable cases\protect\footnotemark}
\caption{General performance of MPPI in 2D and 3D scenarios for both \textit{FOC} and \textit{POC}}
\begin{center}
\small\addtolength{\tabcolsep}{-5.5pt} % resize the table
 \begin{tabular}{|l||l|l||l|l|}
 \hline
 \multirow{2}{*}{\textcolor{blue}{\textbf{Indicators}}} & \multicolumn{2}{c||}{\textcolor{blue}{\textbf{2D Scenario}}} & \multicolumn{2}{c|}{\textcolor{blue}{\textbf{\textbf{3D Scenario}}}}\\ 
 \cline{2-5}
  & \multicolumn{1}{c|}{\textit{FOC}} & \multicolumn{1}{c||}{\textit{POC}} & \multicolumn{1}{c|}{\textit{\textit{FOC}}} & \multicolumn{1}{c|}{\textit{\textit{POC}}}\\ [0.5ex]
 \hline\hline
 \rowcolor{applegreen}
 $N_{\text{col}}(t_{\text{comp}}[\%])$  & 0 (100\%)     & 0 (100\%)     & 0 (100\%)     & 0 (100\%)\\
 \hline
 $t_{\text{av}}$ {[\si{\second}]}           & $122.1\pm1.2$     & $124.5\pm3.0$        & $128.2\pm3.0$        & $129.5\pm1.4$\\ 
 \hline
 $d_{\text{av}}$ {[\si{\metre}]}            & $150.3\pm0.8$     & $151.1\pm1.9$         & $152.9\pm1.8$        & $153.1\pm0.87$\\
 \hline 
 \rowcolor{applegreen}
 $v_{\text{av}}$ {[\si{\metre\per\second}]} & $1.23\pm0.02$      & $1.22\pm0.03$           & $1.19\pm0.02$          & $1.18\pm0.01$\\
 \hline 
 $E_{\text{av}}$ {[\si{\watt\hour}]}        & $6.63\pm0.07$     & $6.76\pm0.16$          & $6.96\pm0.16$         & $7.01\pm0.07$\\ 
 \hline 
 \rowcolor{applegreen}
 $t_{\text{mppi}}$ {[\si{\milli\second}]}   & $18.4\pm\num{0.51}$         & $18.8\pm\num{0.42}$          & $18.2\pm\num{0.21}$  & $18.8\pm\num{0.67}$ \\
 \hline
\end{tabular}
\end{center}
\label{table:Comparison}
\end{table}
%----------------------------------------------------------------------------------------------
In all trials (i.e., 20 trials in total), we can observe that the quadrotor navigates autonomously (i.e., $N_{\text{col}}=0$) while avoiding obstacles with an average flying speed $v_{\text{av}}$ of about \SI{1.2}{\metre\per\second}, which is closer to the maximum specified velocity, i.e., $v_\text{max}=$ \SI{1.5}{\metre\per\second}, regardless of the limited FoV of the sensor in both 2D and 3D scenarios. Note that these 20 trials are equivalent to more than \SI{3}{\kilo\metre} of autonomous navigation. 
Moreover, it can be seen that the average execution time of MPPI per iteration, $t_{\text{mppi}}$, in the 3D scenario is approximately equal to that in the 2D scenario (with very low standard deviation values), showing the superiority of the proposed framework even with 3D-voxel grids and its applicability to be used for 3D navigation tasks. In fact, this is not surprising as the 2D- or 3D-voxel grids are stored and directly used by MPPI on the GPU. 
For other indicators, we can observe that the performance of MPPI in \textit{POC} cases is slightly different than its performance in \textit{FOC} cases, although we have set the same parameters for all simulations. 
The reason behind this difference is that the control variations $\delta \mathbf{u}$ are generated randomly, each time-step $\Delta t$, on the GPU. 
% -------------------------------------- Effect of changing T and v 
\definecolor{LightCyan}{rgb}{0.88,1,1}
\begin{table}[ht!]
\caption{Effect of changing $T$ and $\nu$ on the behavior of the proposed controller}
\begin{center}
\small\addtolength{\tabcolsep}{-4pt} % resize the table
 \begin{tabular}{|l||l|l||l|l|}
 \hline
 \multirow{2}{*}{\hspace{0.2cm}\textcolor{blue}{\textbf{Indicators}}} & \multicolumn{2}{c||}{\textcolor{blue}{\textbf{2D Scenario}}} & \multicolumn{2}{c|}{\textcolor{blue}{\textbf{\textbf{3D Scenario}}}}\\ 
 \cline{2-5}
  & \multicolumn{1}{c|}{\textit{FOC}} & \multicolumn{1}{c||}{\textit{POC}} & \multicolumn{1}{c|}{\textit{\textit{FOC}}} & \multicolumn{1}{c|}{\textit{\textit{POC}}}\\ [0.5ex]
  % ---------------------------------- Case 1 ------------------------------
 \hline\hline
 \rowcolor{LightCyan}
 \multicolumn{5}{|c|}{\textbf{\textit{Tuning Case 1}: \textcolor{red}{$T=75$ (i.e., $t_p = \SI{1.5}{\second}$) and $\nu= 1000$}}}\\
 \hline
\hspace{0.03cm}{\footnotesize\emph{N}\textsubscript{col} (\emph{t}\textsubscript{comp}[\%])}   & \hspace{0.1cm}{\footnotesize 1 (57.4}${\scriptstyle\pm}${\footnotesize 30)}   & \hspace{0.1cm}{\footnotesize 1 (57.4}${\scriptstyle\pm}${\footnotesize 30)}  & \hspace{0.1cm}{\footnotesize 1 (40.3}${\scriptstyle\pm}${\footnotesize 6)}    & \hspace{0.1cm}{\footnotesize 1 (36.3}${\scriptstyle\pm}${\footnotesize 9)}\\
 % ---------------------------------- Case 2 ------------------------------
 \hline\hline
 \rowcolor{LightCyan}
 \multicolumn{5}{|c|}{\textbf{\textit{Tuning Case 2}: \textcolor{red}{$T=100$ (i.e., $t_p = \SI{2}{\second}$) and $\nu= 1000$}}}\\
 \hline
 \hspace{0.03cm}{\footnotesize\emph{N}\textsubscript{col} (\emph{t}\textsubscript{comp}[\%])} & \hspace{0.3cm}{\footnotesize0 (100\%)}     & \hspace{0.3cm}{\footnotesize0 (100\%)}     & \hspace{0.25cm}{\footnotesize0 (100\%)}     & \hspace{0.25cm}{\footnotesize0 (100\%)}\\
 \hline
\hspace{0.05cm}{\footnotesize\emph{d}\textsubscript{av} [m]}           & \hspace{0.2cm}{\footnotesize 51.7}${\scriptstyle\pm}${\footnotesize 0.79}         & \hspace{0.2cm}{\footnotesize 51.7}${\scriptstyle\pm}${\footnotesize 1.1}        & \hspace{0.2cm}{\footnotesize 51.1}${\scriptstyle\pm}${\footnotesize 0.16}        & \hspace{0.2cm}{\footnotesize 52.1}${\scriptstyle\pm}${\footnotesize 0.68}\\
 %$v_{\text{av}}$ {[\si{\metre\per\second}]} & 1.37 & 1.33 & 1.34 & 1.33\\
 % ---------------------------------- Case 3 ------------------------------
 \hline\hline
 \rowcolor{LightCyan}
 \multicolumn{5}{|c|}{\textbf{\textit{Tuning Case 3}: \textcolor{red}{$T=125$ (i.e., $t_p = \SI{2.5}{\second}$) and $\nu= 1000$}}}\\
 \hline
 \hspace{0.03cm}{\footnotesize\emph{N}\textsubscript{col} (\emph{t}\textsubscript{comp}[\%])}  & \hspace{0.3cm}{\footnotesize0 (100\%)}     & \hspace{0.3cm}{\footnotesize0 (100\%)}     & \hspace{0.3cm}{\footnotesize0 (100\%)}     & \hspace{0.3cm}{\footnotesize0 (100\%)}\\
 \hline
 %\rowcolor{green}
\hspace{0.05cm}{\footnotesize\emph{d}\textsubscript{av} [m]}           & \hspace{0.2cm}{\footnotesize\textbf{50.4}}${\scriptstyle\pm}${\footnotesize 0.86}          & \hspace{0.2cm}{\footnotesize\textbf{50.7}}${\scriptstyle\pm}${\footnotesize 0.25}         & \hspace{0.2cm}{\footnotesize\textbf{50.2}}${\scriptstyle\pm}${\footnotesize 0.49}         & \hspace{0.2cm}{\footnotesize\textbf{50.5}}${\scriptstyle\pm}${\footnotesize 0.36}\\
 % ---------------------------------- Case 4 ------------------------------
 \hline\hline
 \rowcolor{LightCyan}
 \multicolumn{5}{|c|}{\textbf{\textit{Tuning Case 4}: \textcolor{red}{$T=150$ (i.e., $t_p = \SI{3}{\second}$) and $\nu= 300$}}}\\
 \hline
 \hspace{0.03cm}{\footnotesize\emph{N}\textsubscript{col} (\emph{t}\textsubscript{comp}[\%])}  & \hspace{0.3cm}{\footnotesize0 (100\%)}     & \hspace{0.3cm}{\footnotesize0 (100\%)}     & \hspace{0.3cm}{\footnotesize0 (100\%)}     & \hspace{0.3cm}{\footnotesize0 (100\%)}\\
 \hline
\hspace{0.05cm}{\footnotesize\emph{d}\textsubscript{av} [m]}           & \hspace{0.2cm}{\footnotesize 52.4}${\scriptstyle\pm}${\footnotesize 0.78}         & \hspace{0.2cm}{\footnotesize 52.5}${\scriptstyle\pm}${\footnotesize 0.36}         & \hspace{0.2cm}{\footnotesize 52.8}${\scriptstyle\pm}${\footnotesize 0.74}        & \hspace{0.2cm}{\footnotesize 53.0}${\scriptstyle\pm}${\footnotesize 0.45}\\
 %$v_{\text{av}}$ {[\si{\metre\per\second}]} & 1.37 & 1.33 & 1.34 & 1.33\\
  % ---------------------------------- Case 5 ------------------------------
 \hline\hline
 \rowcolor{LightCyan}
 \multicolumn{5}{|c|}{\textbf{\textit{Tuning Case 5}: \textcolor{red}{$T=150$ and $\nu= 500$}}}\\
 \hline
 \hspace{0.03cm}{\footnotesize\emph{N}\textsubscript{col} (\emph{t}\textsubscript{comp}[\%])}  & \hspace{0.3cm}{\footnotesize0 (100\%)}     & \hspace{0.3cm}{\footnotesize0 (100\%)}     & \hspace{0.3cm}{\footnotesize0 (100\%)}     & \hspace{0.3cm}{\footnotesize0 (100\%)}\\
 \hline
\hspace{0.05cm}{\footnotesize\emph{d}\textsubscript{av} [m]}           & \hspace{0.2cm}{\footnotesize 51.9}${\scriptstyle\pm}${\footnotesize 0.59}         & \hspace{0.2cm}{\footnotesize 51.5}${\scriptstyle\pm}${\footnotesize 0.85}         & \hspace{0.2cm}{\footnotesize 52.1}${\scriptstyle\pm}${\footnotesize 0.25}        & \hspace{0.2cm}{\footnotesize 52.6}${\scriptstyle\pm}${\footnotesize 0.36}\\
 %$v_{\text{av}}$ {[\si{\metre\per\second}]} & 1.37 & 1.33 & 1.34 & 1.33\\
 % ---------------------------------- Case 6 ------------------------------
 \hline\hline
 \rowcolor{LightCyan}
 \multicolumn{5}{|c|}{\textbf{\textit{Tuning Case 6}: \textcolor{red}{$T=150$ and $\nu= 800$}}}\\
 \hline
 \hspace{0.03cm}{\footnotesize\emph{N}\textsubscript{col} (\emph{t}\textsubscript{comp}[\%])} & \hspace{0.3cm}{\footnotesize0 (100\%)}     & \hspace{0.3cm}{\footnotesize0 (100\%)}     & \hspace{0.3cm}{\footnotesize0 (100\%)}     & \hspace{0.3cm}{\footnotesize0 (100\%)}\\
 \hline
 %\rowcolor{green}
 \hspace{0.05cm}{\footnotesize\emph{d}\textsubscript{av} [m]}            & \hspace{0.2cm}{\footnotesize\textbf{50.9}}${\scriptstyle\pm}${\footnotesize 0.25}          & \hspace{0.2cm}{\footnotesize\textbf{51.0}}${\scriptstyle\pm}${\footnotesize 0.46}         & \hspace{0.2cm}{\footnotesize\textbf{51.5}}${\scriptstyle\pm}${\footnotesize 0.93}         & \hspace{0.2cm}{\footnotesize\textbf{51.9}}${\scriptstyle\pm}${\footnotesize 0.33}\\
 \hline
\end{tabular}
\end{center}
\label{table:Np effects}
\end{table}
%\guillaume{Table I and table II are very difficult to read. Lots of accronyme. I av to go back in the previous page to understand the notation. You have to find a way to give the results in a better way. Maybe recall the notation in the legend?}
The number of timesteps $T$ (which is related to the time horizon $t_p$ and control frequency) and the exploration variance $\nu$ (which is related to the number of sampled trajectories $K$) play an important role in determining the behavior of MPPI. Therefore, we tested six different tuning cases, where different values of $T$ and $\nu$ have been considered, as shown in Table~\ref{table:Np effects}. In the first three tuning cases, we study the influence of changing $T$ where $\nu$ remains constant (i.e., $\nu=1000$ as defined before in Table~\ref{table:SysParameters}). While the effect of changing $\nu$ is studied in the last three cases, where $\nu$ was varied between $300$ and $800$. For the sake of simplicity, each tuning case was tested by letting the quadrotor navigates only to $G_1= (23,38,1.5)$ then lands. In each tuning case, we conducted $3$ trials. As a consequence, the total trials for both \textit{FOC} and \textit{POC} cases are $72$. For \textit{Tuning Case 1}, where a short time horizon $t_p$ is chosen, it can be clearly noticed that the MPPI controller is unable to complete the trials at a satisfactory rate, where the success rate $t_{\text{comp}}$ varies from $36.3\%$ to a maximum of $57.4\%$ in both 2D \textit{FOC} and \textit{POC} (1/3 successful trials). For other tuning cases (namely, from case 2 to 6), we can observe that the controller performs perfectly and is able to successfully complete all trials while avoiding obstacles. We can also notice that as $T$ and $\nu$ increase, the performance of the controller improves. Clearly, for \textit{Tuning Case 3} and \textit{6} where high values of $T$ and $\nu$ are chosen, we can see that the average flying distance $d_{\text{av}}$ (for both 2D and 3D scenarios) is the shortest (compared to other successful cases) with low standard deviation values, which means that the quadrotor is taking a more direct route leading to the goal. However, having too long time horizons increase the computational effort dramatically because each trajectory takes more time to simulate. While, if the exploration variance $\nu$ is too large, the controller produces control inputs with significant chatter. We also notice during our simulations that higher values of control weight matrix $R$ leads to empirically slow convergence of MPPI towards the goal and fluctuated motion.
For this reason, in all experiments, we set $T$, $\nu$, and $R$ to the values given in Table \ref{table:SysParameters}, where MPPI performs very consistently for all given goals. 
\section{Conclusions and future work}\label{sec:conclusion}
Within this work, we proposed an extension to the classical MPPI framework that enables the robot to navigate autonomously in 2D or 3D environments while avoiding collisions with obstacles. The key point of our proposed framework is to provide MPPI with a 2D or 3D grid representing the real world to perform collision-free navigation. This framework has been successfully tested on realistic simulations using quadrotor, considering both fully and partially observable cases. The current simulations illustrate the efficiency and robustness of the proposed controller for 2D and 3D  navigation tasks. 
Although the \textit{theoretical} stability of MPPI has not been addressed and proofed in the literature, its \textit{practical} stability can be achieved by setting the MPPI parameters carefully as we described previously. We will explore the possible methods that may allow in the future to study the \textit{theoretical} stability of MPPI. Our future work will also include the implementation of the framework in practical applications. 

\addtolength{\textheight}{-12cm}   % This command serves to balance the column lengths

%%%%%%%%%%%%%%%%%%%%%%%%%%%%%%%%%%%%%%%%%%%%%%%%%%%%%%%%%%%%%%%%%%%%%%%%%%%%%%%%

%%%%%%%%%%%%%%%%%%%%%%%%%%%%%%%%%%%%%%%%%%%%%%%%%%%%%%%%%%%%%%%%%%%%%%%%%%%%%%%%
\bibliographystyle{IEEEtran}
\bibliography{references}            
%%%%%%%%%%%%%%%%%%%%%%%%%%%%%%%%%%%%%%%%%%%%%%%%%%%%%%%%%%%%%%%%%%%%%%%%%%%%%%%%

\end{document}

%% file: layers.tikz
\begin{tikzpicture}[scale=.62,every node/.style={minimum size=1cm},on grid]
% ---------------- 1st layer ---------------------------------------		
    \begin{scope}[
            yshift=-83,every node/.append style={
            yslant=0.5,xslant=-1},yslant=0.5,xslant=-1
            ]
        % opacity to prevent graphical interference
        \fill[green!50,fill opacity=0.9] (0,0) rectangle (5,5);
        \draw[step=5mm, black] (0,0) grid (5,5); %defining grids
        \draw[black,very thick] (0,0) rectangle (5,5); %marking borders
        \fill[black] (0.55,0.55) rectangle (0.95,0.95);
        \fill[black] (0.05,1.55) rectangle (0.45,1.95);
         \fill[black] (0.05,3.05) rectangle (0.45,4.45);
        \fill[black] (1.05,2.05) rectangle (1.45,2.45);
        \fill[black] (1.05, 0.05) rectangle (2.45,0.45);
        \fill[black] (3.05, 1.05) rectangle (3.45,1.45);
        \fill[blue] (3.05, 0.05) rectangle (3.45,0.45); % Unknown cell
        \fill[blue] (2.55, 1.05) rectangle (2.95,1.45); % Unknown cell
        \fill[blue] (0.05, 2.05) rectangle (0.45,2.45); % Unknown cell
        \fill[black] (1.55, 1.05) rectangle (1.95,1.45);
        \fill[black] (2.05,2.05) rectangle (2.45,2.45); % center pixel
        \fill[black] (2.05,3.05) rectangle (2.45,3.45);
        %\draw[-latex,thick](2,-1)node[right, rotate=0]{\textbf{$1^{st}$ Layer} $\ell_1$} to[out=180,in=270] (2.5,0);
    \end{scope}
% ------------------- 2nd layer ---------------------------------    	
    \begin{scope}[
    	yshift=-63,every node/.append style={
    	    yslant=0.5,xslant=-1},yslant=0.5,xslant=-1
    	             ]
        \fill[green!50,fill opacity=.9] (0,0) rectangle (5,5);
        \fill[yellow,fill opacity=.9] (0,0) rectangle (1.5,1.5); % Mask
        \draw[black,very thick] (0,0) rectangle (5,5);
        \draw[step=5mm, black] (0,0) grid (5,5);
        \fill[black] (0.55,0.55) rectangle (0.95,0.95);
        \fill[black] (0.05,1.55) rectangle (0.45,1.95);
        \fill[black] (0.05,3.05) rectangle (0.45,4.45);
        \fill[black] (1.05,2.05) rectangle (1.45,2.45);
        \fill[black] (1.05, 0.05) rectangle (2.45,0.45);
        \fill[black] (3.05, 1.05) rectangle (3.45,1.45);
        \fill[blue] (3.05, 0.05) rectangle (3.45,0.45); % Unknown cell
        \fill[blue] (2.55, 1.05) rectangle (2.95,1.45); % Unknown cell
        \fill[blue] (0.05, 2.05) rectangle (0.45,2.45); % Unknown cell
        \fill[black] (1.55, 1.05) rectangle (1.95,1.45);
        \fill[black] (2.05,2.05) rectangle (2.45,2.45); % center pixel
        \fill[black] (2.05,3.05) rectangle (2.45,3.45);
    \end{scope}
    % ----------------------------3nd layer ----------------------    	
    \begin{scope}[
    	yshift=-43,every node/.append style={
    	    yslant=0.5,xslant=-1},yslant=0.5,xslant=-1
    	             ]
        \fill[green!50,fill opacity=.9] (0,0) rectangle (5,5);
        \fill[yellow,fill opacity=.9] (0,0) rectangle (1.5,1.5); % Mask
        \draw[black,very thick] (0,0) rectangle (5,5);
        \draw[step=5mm, black] (0,0) grid (5,5);
        \fill[black] (0.55,0.55) rectangle (0.95,0.95);
        \fill[black] (0.05,1.55) rectangle (0.45,1.95);
        \fill[black] (0.05,3.05) rectangle (0.45,4.45);
        \fill[black] (1.05,2.05) rectangle (1.45,2.45);
        \fill[black] (1.05, 0.05) rectangle (2.45,0.45);
        \fill[black] (3.05, 1.05) rectangle (3.45,1.45);
        \fill[blue] (3.05, 0.05) rectangle (3.45,0.45); % Unknown cell
        \fill[blue] (2.55, 1.05) rectangle (2.95,1.45); % Unknown cell
        \fill[blue] (0.05, 2.05) rectangle (0.45,2.45); % Unknown cell
        \fill[black] (1.55, 1.05) rectangle (1.95,1.45);
        \fill[black] (2.05,2.05) rectangle (2.45,2.45); % center pixel
        \fill[black] (2.05,3.05) rectangle (2.45,3.45);
        \fill[red] (0.06,0.06) circle (.15); % robot position
    \end{scope}
 % ----------------------------4nd layer ----------------------    	
    \begin{scope}[
    	yshift=-23,every node/.append style={
    	    yslant=0.5,xslant=-1},yslant=0.5,xslant=-1
    	             ]
        \fill[green!50,fill opacity=.9] (0,0) rectangle (5,5);
        \fill[yellow,fill opacity=.9] (0,0) rectangle (1.5,1.5); % Mask 
        \draw[black,very thick] (0,0) rectangle (5,5);
        \draw[step=5mm, black] (0,0) grid (5,5);
        \fill[black] (0.55,0.55) rectangle (0.95,0.95);
        \fill[black] (0.05,1.55) rectangle (0.45,1.95);
        \fill[black] (0.05,3.05) rectangle (0.45,4.45);
        \fill[black] (1.05,2.05) rectangle (1.45,2.45);
        \fill[black] (1.05, 0.05) rectangle (2.45,0.45);
        \fill[black] (3.05, 1.05) rectangle (3.45,1.45);
        \fill[blue] (3.05, 0.05) rectangle (3.45,0.45); % Unknown cell
        \fill[blue] (2.55, 1.05) rectangle (2.95,1.45); % Unknown cell
        \fill[blue] (0.05, 2.05) rectangle (0.45,2.45); % Unknown cell
        \fill[black] (1.55, 1.05) rectangle (1.95,1.45);
        \fill[black] (2.05,2.05) rectangle (2.45,2.45); % center pixel
        \fill[black] (2.05,3.05) rectangle (2.45,3.45);
    \end{scope} 
% ------------------- layer N ------------------------------------    
    \begin{scope}[
    	yshift=23,every node/.append style={
    	yslant=0.5,xslant=-1},yslant=0.5,xslant=-1
    	             ]
    	\fill[green!50,fill opacity=.9] (0,0) rectangle (5,5);
    	\draw[step=5mm, black] (0,0) grid (5,5);
    	\draw[black,very thick] (0,0) rectangle (5,5);
    	\draw[black,dashed] (0,0) rectangle (5,5);
    	\fill[black] (0.55,0.55) rectangle (0.95,0.95);
        \fill[black] (0.05,1.55) rectangle (0.45,1.95);
        \fill[black] (0.05,3.05) rectangle (0.45,4.45);
        \fill[black] (1.05,2.05) rectangle (1.45,2.45);
        \fill[black] (1.05, 0.05) rectangle (2.45,0.45);
        \fill[black] (3.05, 1.05) rectangle (3.45,1.45);
        \fill[blue] (3.05, 0.05) rectangle (3.45,0.45); % Unknown cell
        \fill[blue] (2.55, 1.05) rectangle (2.95,1.45); % Unknown cell
        \fill[blue] (0.05, 2.05) rectangle (0.45,2.45); % Unknown cell
        \fill[black] (1.55, 1.05) rectangle (1.95,1.45);
        \fill[black] (2.05,2.05) rectangle (2.45,2.45); % center pixel
        \fill[blue] (4.05,3.05) rectangle (4.45,3.45);
        \fill[black] (4.05,2.05) rectangle (4.45,2.45);
        \fill[black] (2.05,3.05) rectangle (3.45,3.45);
        \fill[black] (2.05,4.05) rectangle (2.95,4.45);
        \fill[black] (4.05, 4.55) rectangle (4.45,4.95);
    \end{scope}
    %end of drawing grids

    %-----------putting arrows and labels---------------:
    \draw[-latex,thick](5.8,-0.25)node[right]{$1^\text{st}$ Layer $\ell_1$}
        to[out=150,in=50] (4.7,-0.25);
    \draw[-latex,thick] (5.8,0.45) node[right]{$2^{\text{nd}}$ Layer $\ell_2$}
         to[out=150,in=50] (4.7,0.45); 
      \draw[-latex,thick] (5.8,1.15) node[right]{$3^{\text{rd}}$ Layer $\ell_3$}
         to[out=150,in=50] (4.7,1.15); 
     \draw[-latex,thick] (5.8,1.85) node[right]{$4^{\text{th}}$ Layer $\ell_4$}
         to[out=150,in=50] (4.7,1.85);
    \draw[-latex,thick] (5.6,2.60) node[right]{$N^{\text{th}}$ Layer $\ell_N$}
         to[out=200,in=-70] (4.5,3.0); 
     %\draw[-latex,thick](5.8,3.90)node[right]{$2^{\text{nd}}$ Layer $\ell_N$}
      %  to[out=180,in=60] (4.7,3.90);     
    % Unknown cell
    \draw[-latex,thick,red](5.5,-1.15)node[right]{\textcolor{black}{$\mathcal{M}_{\text{unk}}\equiv-1$}}  to[out=180,in=90] (3.0,-1.2);    
     % Occ cell
    \draw[-latex,thick,red](4.5,-1.9)node[right]{\textcolor{black}{$\mathcal{M}_{\text{occ}}\equiv 1$}} to[out=180,in=90] (1.5,-1.9); 
 % Free cell
    \draw[-latex,thick,red](3.5,-2.6)node[right]{\textcolor{black}{$\mathcal{M}_{\text{free}}\equiv 0$}} to[out=180,in=90] (0.5,-2.5); % (0.5,-2.4) for arrow pointing to the cell
    
    % 3D Mask and robot pose
     \draw[-latex,thick,red] (2.0,5.8) node[right]{3D $\mathcal{F}_{fov}$, Dim: $(3 \times 3 \times 3)$,} to[out=150,in=70] (-0.95,-0.1);
     \draw[thick,red] (2.0,5.1) node[right]{where robot is placed in $\ell_3$,};
     \draw[thick,red] (3.1,4.4) node[right]{and $\ell_2:\ell_4$ are updated.};
      \draw[-latex,thick,red] (2.0,5.8) to[out=150,in=70] (-0.5,-1.0);
      \draw[-latex,thick,red] (2.0,5.8) to[out=150,in=70] (0.3,-1.9);
      % Robot Pose
      %\draw[-latex,thick,red](-5.4,-1.1)node[right, rotate = -27 ]{Robot Located at $(0,0,\ell_3)$} to[out=50,in=190] (-0.2,-1.45);
      \draw[-latex,thick,red](-4.6,-2.1)node[right]{Robot} to[out=50,in=190] (-0.2,-1.45);
      \draw[thick,red](-4.6,-2.8) node[right]{at $(0,0,\ell_3)$};
\end{tikzpicture}